# Robotic Mapping with Polygonal Random Fields

**Mark A. Paskin**
Computer Science Department
Stanford University
Stanford, California 94305

**Sebastian Thrun**
Computer Science Department
Stanford University
Stanford, California 94305

## Abstract

Two types of probabilistic maps are popular in the mobile robotics literature: occupancy grids and geometric maps. Occupancy grids have the advantages of simplicity and speed, but they represent only a restricted class of maps and they make incorrect independence assumptions. On the other hand, current geometric approaches, which characterize the environment by features such as line segments, can represent complex environments compactly. However, they do not reason explicitly about occupancy, a necessity for motion planning; and, they lack a complete probability model over environmental structures. In this paper we present a probabilistic mapping technique based on polygonal random fields (PRF), which combines the advantages of both approaches. Our approach explicitly represents occupancy using a geometric representation, and it is based upon a consistent probability distribution over environments which avoids the incorrect independence assumptions made by occupancy grids. We show how sampling techniques for PRFs can be applied to localized laser and sonar data, and we demonstrate significant improvements in mapping performance over occupancy grids.

## 1 INTRODUCTION

Probabilistic reasoning has led to significant advances in robotic mapping in the past decade (Thrun, 2002). A few different types of probabilistic map representations have been proposed, but by far the most popular is the occupancy grid (Elfes, 1989; Moravec, 1988). Occupancy grids have been quite successful in mapping with laser rangefinder data, but they make poor approximations that limit their applicability. As a raster representation, occupancy grids represent structure at a single scale, and they introduce discretization error. More importantly, occupancy grids make independence assumptions that are clearly incorrect: they assume all grid cells are independent of each other, in both the prior and posterior distributions. When used with laser data, these problems are easily masked by the volume and high quality of the data; but when applied to more ambiguous data, such as sonar, these problems result in maps of poor quality.

The classical literature on robot motion planning routinely assumes geometric maps with polygonal boundaries between free and occupied space (Latombe, 1991). Such a representation avoids the discretization errors introduced by occupancy grids, and it permits compact representations of complex environments, with structures at arbitrary scales. This has led to a number of approaches to inferring geometric maps from range data. For example, Tardós *et al.* (2002) have used the Hough transform to compute line segment and corner features from sonar data for a Kalman filter mapping technique. Veeck and Burgard (2004) present an optimization technique for fitting polylines to range data (as well as a good survey on other such approaches). However, these techniques reason only about boundaries, not occupancy, which is crucial for motion planning; in addition, they lack a coherent probabilistic model over possible maps.

In this paper, we investigate how probabilistic models of occupancy can be combined with geometric representations. Remarkably, there exists a family of probability distributions—called *polygonal random fields* (PRF)—over representations of this sort. The theoretical properties of PRFs have been studied in the statistics literature, but to our knowledge, PRFs have not yet been applied to real data. In this paper we apply them to robotic mapping problems with laser and sonar data. We show how PRFs can support the complex reasoning required to interpret sonar data, which is notoriously difficult. Our experimental results demonstrate that PRFs can yield significantly better maps than occupancy grids, especially on sonar data. In addition, where the data are ambiguous, PRFs exhibit extrapolation that could prove useful for informed exploration.

Below we present the necessary background on robotic mapping, occupancy grids, and polygonal random fields. In Section 2 we describe how, using sampling techniques, PRFs can be used to solve robotic mapping problems with laser and sonar data. Section 3 presents our experimental results, and Section 4 concludes.

## 1.1 ROBOTIC MAPPING

We consider a simple mapping problem where a mobile robot builds a representation of the free and occupied space in a static, two-dimensional environment (Thrun, 2002). For simplicity, we assume the localization problem has been solved, i.e., the robot's poses are known. These mapping problems are often addressed with laser or sonar rangefinder sensors, which emit a light or sound wave and sense its reflection off of the nearest object. In clean readings, where the wave is returned directly to the sensor, the relative error in both rangefinding technologies is around 1%.

Sonar arrays are much cheaper than laser rangefinders, but they have several disadvantages. Sonar arrays give fewer range readings at a lower rate; they have a smaller effective range; and they are susceptible to multi-path effects. The most important disadvantage, however, is that the spread angle of a sonar rangefinder is typically $20°$, whereas a laser beam's spread is less than $1°$. This makes sonar observations harder to interpret than laser observations: while a laser observation effectively indicates a point of contact with an object, a sonar observation yields an arc that *contains* a point of contact. Examples of laser and sonar range data are shown in Figures 4(d) and 5(a).

## 1.2 OCCUPANCY GRIDS

The most successful technique for robotic mapping with laser and sonar data involves occupancy grids (Elfes, 1989). The environment is divided into a grid of cells (typically 3–10 cm across for indoor environments), and each cell is associated with a binary random variable representing whether the cell is occupied by some object. Assuming that all cells and observations are independent permits the posterior for each cell to be computed from the observations using a simple update rule (Moravec, 1988). Rather than using the natural sensor model of the range measurement $r$ given the robot's pose $x$ and the map $m$, $\Pr\{r \mid x, m\}$, this rule uses an *inverse sensor model* $\Pr\{m_{ij} \mid x, r\}$ which predicts the occupancy of a single cell $m_{ij}$ given the robot's pose and the range measurement. For example, if a laser beam passes through a cell, the cell is likely free; if a cell lies in the impact arc of a sonar observation, it is more likely occupied. Thus, each cell is independently updated with each observation. Typical occupancy grids obtained from laser and sonar data are shown in Figures 4(e) and 5(b).

Occupancy grids are a popular mapping technique, but they suffer from some problems. The first is the restricted hypothesis space imposed by the grid representation; it introduces discretization error and limits the scale at which the environment can be modeled. More importantly, grids make it difficult to interpret sonar observations. A sonar sensor is more likely to receive a return from a surface that is oriented perpendicular to the sound wave's direction of travel, because the sound wave will be reflected back to the sensor. But this reasoning is difficult in occupancy grids, which have surfaces at only two orientations.

A more serious problem is the independence assumptions required to make occupancy grids efficient. The assumption that the grid cells are independent of one another is clearly incorrect: if a given cell is occupied, then it is much more likely that an adjacent cell is also occupied. And, the assumption that grid cells are independent *given observations* is ridiculous: if a laser beam travels unimpeded through a given cell $M_{ij}$, it is almost certain that it also passed through the cells on the segment from the sensor to $M_{ij}$; if a cell intersecting the impact arc of a sonar observation is occupied, the other cells on the arc are less likely occupied.[1]

In spite of these problems, occupancy grids are routinely and successfully applied to laser data. This is possible because the incorrect independence assumptions in the prior are quickly overwhelmed by the tremendous volume of low-noise observations generated by a laser rangefinder; indeed, the occupancy grid in Figure 4(e) is not terribly different from the raw data in 4(d). However, the shortcomings of occupancy grids reveal themselves plainly with sonar data. The occupancy grid in Figure 5(b) has several missing walls because sonar readings that do not return directly to the sensor are interpreted as indications of free space.

---

[1] Correcting these problems may seem as simple as choosing the right graphical model—perhaps a grid Markov random field could correct the prior. However, the posterior is much more difficult. A likelihood function which correctly expresses the dependencies between the many cells crossed by a laser beam is complex to express. And reasoning with such models will require sampling or variational approximations. We are unaware of work along these lines, but see (Berler and Shimony, 1997) for a related approach that obtains modest improvements, though with significant computational cost.

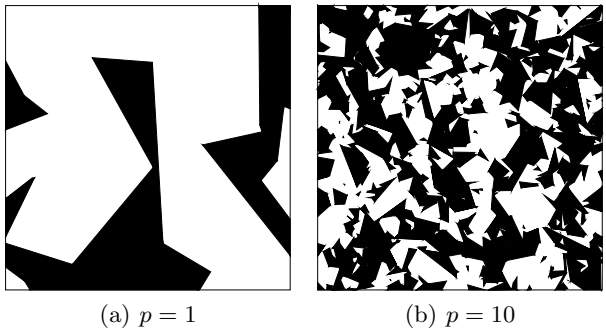

(a) $p = 1$      (b) $p = 10$

Figure 1: Two samples from the Arak process for varying scale parameter $p$ (in the unit square).

And, the remaining walls are indistinct because every cell on the sonar arc is treated as occupied.

## 1.3 POLYGONAL RANDOM FIELDS

In this paper we investigate the advantages of robotic mapping with a richer class of maps and more sophisticated priors, called polygonal random fields. Below we present a brief introduction to these distributions.

### 1.3.1 Polygonal colorings

A *coloring* is a function that classifies all points in the plane $\Re^2$ as either black or white; a *polygonal coloring* has the additional property that the color discontinuities form line segments. While they color the entire plane, we typically reason about polygonal colorings through a finite, convex *window*; for simplicity we will assume the window to be an axis-aligned rectangle. Figure 1 shows two examples of polygonal colorings, viewed through a unit square window.

A polygonal coloring $\chi$ can be represented compactly by two pieces of information. The first is a *discontinuity graph*, $\gamma$, whose edges represent the boundaries between black and white regions. Each vertex of $\gamma$ has a location in the window's interior or on its boundary. Valid discontinuity graphs satisfy three properties: each boundary vertex is connected to exactly one other vertex; each interior vertex is connected to exactly two other vertices; and none of the edges cross. We can represent a coloring $\chi$ by its discontinuity graph $\gamma$ and the color of a single *anchor point* $p_0$ inside the window: the color of any other point $q$ in the window can be inferred by tracing a path from $p_0$ to $q$ and counting the number of crossed edges.

Polygonal colorings are obviously better suited to represent the environments of mobile robots than occupancy grids. They are vector representations, as opposed to occupancy grids, which are raster representations; thus, they avoid discretization error, can represent structure at arbitrary scales, and can represent simple environments far more compactly. And, they explicitly represent the orientations of object surfaces, which is useful for interpreting sonar data.

### 1.3.2 The Arak Process

Polygonal colorings are complicated structures: they have discrete and continuous components; they have varying and unbounded dimensionality; and, they obey complex geometric constraints. It is astounding, then, that there exists a consistent family of probability distributions over polygonal colorings. It was first discovered by Arak (Arak, 1982), and was later generalized to permit more than two colors (Arak and Surgailis, 1989) and more complicated graph structures (Arak *et al.*, 1993). Collectively, these probability distributions are called *polygonal random fields (PRF)*. In this paper we use the simplest PRF, the *Arak process*.

There are a few different ways to characterize the Arak process. It can be defined in terms of a Poisson line process, or by the evolution of a dynamic particle system. For our purposes the definition given in (Clifford and Nicholls, 1994), which characterizes the Arak process in terms of discontinuity graphs, is the most useful. Because polygonal colorings are complicated objects, we must first define a measure over polygonal colorings before defining the distribution. The fundamental measure for polygonal colorings is defined by

$$\lambda(\mathrm{d}\chi) \triangleq p^{|E|} \prod_{e \in E} |e|^{-1} \prod_{v \in V} \sin(\phi_v) \nu(\mathrm{d}\chi) \qquad (1)$$

where $p$ is the *scale parameter*, $V$ and $E$ are the vertices and edges of $\chi$'s discontinuity graph, $|e|$ is the length of the edge $e$, and $\phi_v$ is the *associated angle* of vertex $v$: if $v$ is an interior vertex, then this is the smaller of the two angles made by its two incident edges; if $v$ is on the window's boundary, then this is the smaller of the two angles made by $v$'s only edge with the boundary tangent. $\nu(\mathrm{d}\chi)$ is an elementary base measure over colorings built from Poisson point processes and counting measures (see (Clifford and Nicholls, 1994) for details); we can safely ignore it because in the inference algorithm it is a constant.

Using this measure, the Arak distribution can then be expressed by the Gibbs distribution

$$Q(\mathrm{d}\chi) = \frac{1}{Z} e^{-F(\chi)} \lambda(\mathrm{d}\chi) \qquad (2)$$

where $F(\chi)$ is a non-negative *potential function* over the coloring. To get the Arak distribution, we choose

$$F_0(\chi) \triangleq 2p \sum_{e \in E} |e| \qquad (3)$$

The colorings in Figure 1 are samples from this distribution for two values of the scale parameter $p$.

From the form of the Arak distribution we can infer several nice properties. First, the distribution prefers simpler colorings: for each additional edge introduced, the measure (1) charges us a multiplicative factor of $p$. Simpler color boundaries are also preferred: both the measure (1) and the potential (3) make long edges less likely, and the measure (1) prefers joint angles to be close to $90°$. The distribution is sensitive only to the discontinuity set and not the actual colors, so the prior probability a point is black is 50%.

The Arak distribution has many other important traits that are of theoretical and practical interest. It is isotropic (the likelihood of a coloring remains the same under transformations that preserve lengths); it is stationary (the distribution associated with a window remains the same, no matter where it is in the plane); it is consistent (the distribution agrees with itself on any subwindow); and it is solvable (the normalization constant $Z$ is a simple function of the scale parameter $p$ and the window). In addition, several statistics can be computed using simple formulas; e.g., the expected number of edges in the unit square is $4p + 4\pi p^2$. Perhaps the most striking property is a continuous form of the Markov property on graphs: the coloring inside a convex region $R$ is conditionally independent of the coloring outside $R$ given the coloring in an $\varepsilon$-band around its boundary.

Unlike occupancy grids, the Arak process has significant spatial correlation. If two points are at distance $d$, then with probability $\frac{1}{2}(1 + \exp(-4pd))$ they have the same color.[2] Nearby points are likely to have the same color, and as the distance between them grows, the probability converges to chance.

In this paper we are concerned with inferring a posterior distribution over polygonal colorings given data. As is the case with any Gibbs distribution, we can incorporate the evidence by updating the potential. Given observed data $D$ and a likelihood model $\Pr\{D \mid \chi\}$, we can update our potential to

$$F_D(\chi) \triangleq F_0(\chi) - \log \Pr\{D \mid \chi\} \qquad (4)$$

The distribution (2) then represents a posterior distribution over colorings given the evidence $D$. For example, imagine a data set of real numbers $x_i$, each associated with a location $q_i$ in the plane. Assume that $X_i$ is generated independently from a Gaussian distribution whose mean $\mu_{\chi(q_i)}$ depends upon the true color at $q_i$. Then the posterior potential is

$$F_D(\chi) \triangleq F_0(\chi) + \sum_i (x_i - \mu_{\chi(q_i)})^2 / 2\sigma^2, \qquad (5)$$

ignoring irrelevant constants.

### 1.3.3 Inference via sampling

Because of the complexity of the space of polygonal colorings, sampling techniques seem the best fit for inference in polygonal random fields—in particular, Markov chain Monte Carlo (MCMC) (Robert and Casella, 2004) is a natural choice. A few such PRF samplers have been proposed; by far the most promising of these is the algorithm proposed in (Clifford and Nicholls, 1994), which is well-suited for efficient implementation. Our implementation is based upon this algorithm, which we refer to as the *C&N sampler*.

The C&N sampler is a Metropolis–Hastings MCMC algorithm. In each step $t$ of the algorithm, we have a current coloring $\chi_t$. (We initialize $\chi_0$ to the all-white coloring.) We sample a candidate new state $\hat{\chi}_{t+1}$ given the current state $\chi_t$ from a proposal distribution $\pi(d\hat{\chi}_{t+1} \mid \chi_t)$, and compute the acceptance probability:

$$\alpha(\hat{\chi}_{t+1} \mid \chi_t) \triangleq \min\left(1, \frac{Q(d\hat{\chi}_{t+1})}{Q(d\chi_t)} \times \frac{\pi(d\chi_t \mid \hat{\chi}_{t+1})}{\pi(d\hat{\chi}_{t+1} \mid \chi_t)}\right)$$

Then we choose the state of the Markov chain so that

$$\chi_{t+1} = \begin{cases} \hat{\chi}_{t+1} & \text{with probability } \alpha(\hat{\chi}_{t+1} \mid \chi_t) \\ \chi_t & \text{with the remaining probability} \end{cases}$$

The proposal distribution is proved to yield a reversible, Harris recurrent Markov chain; this guarantees that in the limit, the states visited by the Markov chain are distributed according to the PRF $Q$.[3]

The C&N proposal distribution consists of several types of moves; see Figure 2. To sample a new coloring, the type of move is sampled, and then the actual move is sampled given the move type and the current coloring. For example, we may sample the interior triangle birth type, sample three vertex locations in the window's interior, and propose adding a new triangle to the discontinuity graph, as depicted in Figure 2(a); if the anchor point $p_0$ is inside the triangle, this move reverses its color. Thus, the net effect of this move is to reverse the color of all points inside the triangle; the color of all other points remains unchanged.

---

[2] To our knowledge this is a new result. It follows from the fact that the number of intersections of the discontinuity graph with a line segment of length $d$ is Poisson distributed with mean $2pd$ (Arak *et al.*, 1993).

[3] Readers familiar with MCMC in variable dimension state spaces may wonder if a "reversible jump" formulation, complete with projective transformations and Jacobians, is necessary. It is not: Clifford and Nicholls give simple conditions that guarantee their moves between states with different dimension satisfy detailed balance.

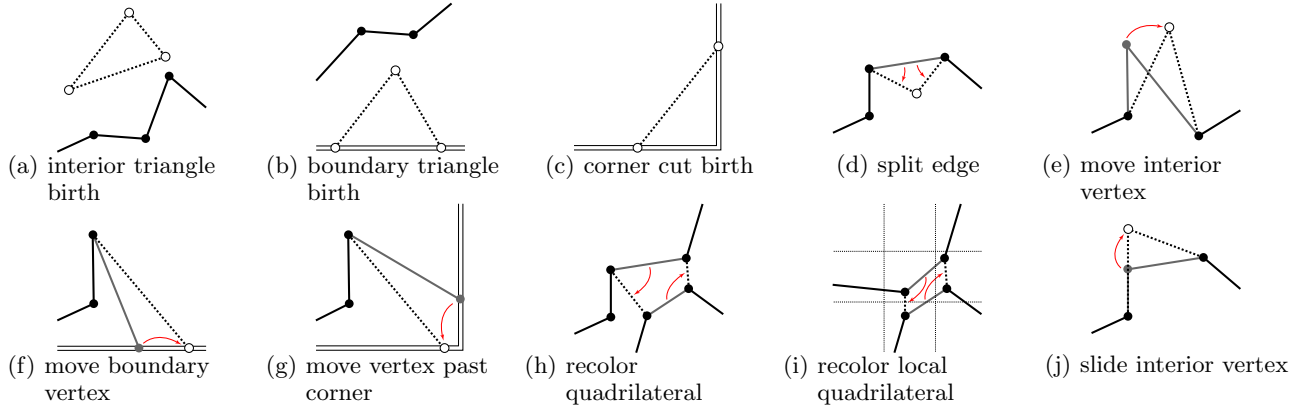

| | | | | |
|---|---|---|---|---|
| (a) interior triangle birth | (b) boundary triangle birth | (c) corner cut birth | (d) split edge | (e) move interior vertex |
| (f) move boundary vertex | (g) move vertex past corner | (h) recolor quadrilateral | (i) recolor local quadrilateral | (j) slide interior vertex |

Figure 2: The proposal moves; new vertices are white and new edges are dashed; old vertices and edges are gray.

Figures 2(a)–2(d) illustrate the move types that introduce new structural elements; Figures 2(e)–2(g) illustrate the move types that adjust the locations of current vertices. One of the most important moves, shown in Figure 2(h), recolors a quadrilateral region by exchanging two current edges for two new edges; this move allows polygons to be merged together or split apart. All of these moves are reversible: the moves that adjust vertex locations can reverse themselves, as can the recolor quadrilateral move; the remaining moves have explicit reverses that are not shown (e.g., there is an interior triangle death move).

The move types of the C&N proposal generate valid colorings with one exception: they can propose to update the discontinuity graph so it has crossing edges, which yields an invalid coloring. In this case, the move must be rejected. To efficiently check for crossing edges, the C&N algorithm makes use of a grid index over the edges of the coloring. Each cell of the grid contains a list of the edges that cross it; to check if a newly proposed edge crosses any existing edges of the coloring, the grid cells that intersect the new edge are visited—this can done quickly using Bresenham's algorithm—and the new edge is tested for intersections only with the edges indexed in those cells. We found this simple strategy surprisingly efficient.

The second half of the Metropolis–Hastings update is to evaluate the likelihood of the proposed state under the PRF. The likelihood of the proposal under the Arak prior can be computed in constant time; all we must do is cache certain statistics of the coloring like the sum of the edge lengths. To estimate the likelihood given data, however, more work is required. A naive implementation would recompute the likelihood of each data point from scratch; a much faster alternative is possible because each move updates a small (triangular or quadrilateral) portion of the coloring. By indexing the data and recomputing the likelihoods of only those data that are sensitive to the updated region, significant speed improvements are possible.

For example, to condition on Gaussian point observations of the sort described above, Clifford and Nicholls (1994) use a grid index to identify those points whose log likelihood must be recomputed in (5).

The C&N sampler can be used to efficiently estimate the expectations of many quantities of interest. For example, we can estimate the probability a particular point is colored black by computing the fraction of samples in which it was black. To generate a visualization of the posterior, we can estimate these probabilities for a grid of points and then display them as an image. To generate a representation like an occupancy grid, we can estimate the fraction of samples in which each square in a grid was completely white. As in recomputing likelihoods, the locality of updates can be exploited to compute these estimates efficiently.

## 2 ROBOTIC MAPPING WITH POLYGONAL RANDOM FIELDS

To our knowledge, neither the Arak process nor the C&N sampler have yet been applied to real data. Below we describe how we applied them to robotic mapping problems. We begin by presenting some modifications to the C&N algorithm that greatly improve its performance on rangefinder data. Then we describe our likelihood models for laser and sonar, as well as indexing strategies to minimize likelihood computations.

### 2.1 ADAPTING THE C&N ALGORITHM

To apply the C&N sampler to robotic mapping, a few modifications were necessary. These enhancements were needed for two reasons: to scale the algorithm to larger, more complex colorings; and, to explore different interpretations of rangefinder data efficiently. Both issues were addressed by introducing new move types into the proposal distribution. It is easy to show these moves satisfy the conditions for detailed balance given in (Clifford and Nicholls, 1994).

The experiments in (Clifford and Nicholls, 1994) apply the C&N sampler to synthetic problems where the colorings have around 100 edges. In our experiments, we often have maps with over 1000 edges. Scaling to colorings of this complexity revealed a problem with the C&N proposal: the recolor move of Figure 2(h)—perhaps the most important move for exploring alternative structures—almost always fails in large colorings. This is because the C&N proposal samples a recolor move by choosing the two edges to replace *uniformly* from the coloring; edges sampled in this way are rarely close to each other, which means the replacement edges frequently cross other edges, yielding an invalid move. To fix this problem we introduced a local version of the recolor move, shown in Figure 2(i). To sample a local recolor move, we use the same edge index used to check for crossing edges: we first sample an edge $e$ uniformly from the coloring, and then sample the second edge uniformly from the set of edges that co-occur with $e$ in some cell of the grid index. This local version of the move had a significantly improved acceptance ratio.

Whereas (Clifford and Nicholls, 1994) considered data that are sensitive to the colors at point locations, rangefinder data are also sensitive to the locations and orientations of color discontinuities. This led to a problem with the C&N move that relocates vertices, shown in Figure 2(e). When moving a vertex, this move also changes the orientations of the incident edges; such changes can drastically reduce the likelihood of rangefinder data, which causes heavy move rejection. To solve this problem, we introduced another move type to relocate vertices: the slide move of Figure 2(j). To sample a slide move, we first sample an interior vertex $v$ uniformly from the coloring, and then sample one of its two incident edges $e$ uniformly. Finally, we sample a new location for $v$ uniformly on the line segment $[v - \frac{1}{2}\vec{e}, v + \vec{e}]$, where $\vec{e}$ is the vector from $v$ to other vertex incident to $e$. (These endpoints were chosen to ensure reversibility.) Thus, the orientation of $e$ does not change. We found that this move can improve the acceptance ratio significantly.

### 2.2 LASER RANGEFINDER DATA

The sensor model we use for laser data is fairly standard. The likelihood of a range reading given the coloring is zero if the sensor's location is in occupied space. Otherwise, it depends only on the distance from the sensor location to the nearest edge along the direction of the beam's travel (which corresponds to the first obstacle). To compute this distance efficiently, we use a line walk in the coloring's edge index. Once the distance to the nearest obstacle is computed, we compute the likelihood of the range observation using a standard likelihood model that mixes a Gaussian measurement error term with a uniform outlier model and some probability of a max-range reading.

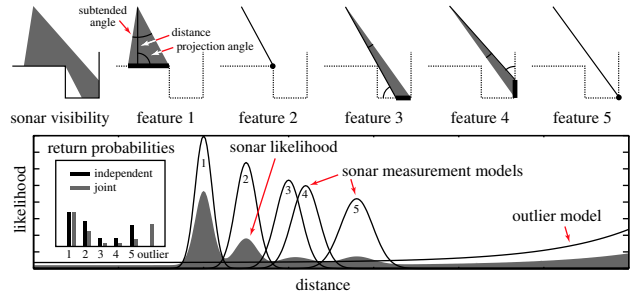

Figure 3: The sonar model, including the extracted features, return probabilities, and likelihood function.

As described in Section 1.3.3, the main cost of the sampling algorithm is in re-evaluating the likelihood of the data; thus, identifying which data are sensitive to a proposed update is essential. We use another grid index over the laser data for this purpose: each cell has a list of the laser measurements that are sensitive to the coloring in the cell. When the coloring is updated, we recompute the likelihoods for only those laser measurements that are indexed in cells overlapping the update region. Because the likelihood function is insensitive to structure beyond the first impact point, we also re-index each observation (on the segment from the sensor to the current impact point) when we recompute its likelihood. We found that this *dynamic indexing* strategy yields substantial speedups.

### 2.3 SONAR RANGEFINDER DATA

One of the strengths of polygonal maps over other representations is that they permit rich, sophisticated sensor models; we exploited this in designing our sonar model. The first step in evaluating a sonar likelihood is to identify which features of the environment may have caused the sonar return. In polygonal models there are two natural types of features: planar surfaces and corners (which are a frequent source of sonar returns). Using our geometric representation we can efficiently identify which faces and corners could have generated the return. We first use the coloring's edge index to identify the edges that overlap the sonar cone (up to its maximum range). Then we use a radial scan line algorithm (Lee, 1978) to compute the segments and corners that are visible to the sonar sensor in $O(k \log k)$ time. Note that only a part of a face may visible to the sensor; an example is given in Figure 3.

Next we compute the probability that each feature generated the sonar return. We use a logistic *independent return model* to compute the probability a feature would generate a sonar return given the pose of the sensor, in the absence of any other features. For corner features, this logistic model depends only upon the distance to the corner. For segment features, it depends upon the closest point on the segment, the projection angle, and the angle subtended in the sensor's field of view (illustrated in Figure 3); a return becomes less likely as the projection distance increases, the projection angle decreases, or the subtended angle decreases. Using these independent return probabilities, we compute the *return probability* of each feature: the probability the return is generated by each feature *in the presence of the other features.* We sort the features by depth, and then compute the return probability for each feature using a construction like that of the Bernoulli distribution: if $q_f$ is the independent return probability of feature $f$, then its return probability is $r_f \triangleq q_f \prod_{g<f}(1-r_g)$. Thus, each feature can generate the return, but only if none of the closer features did.[4]

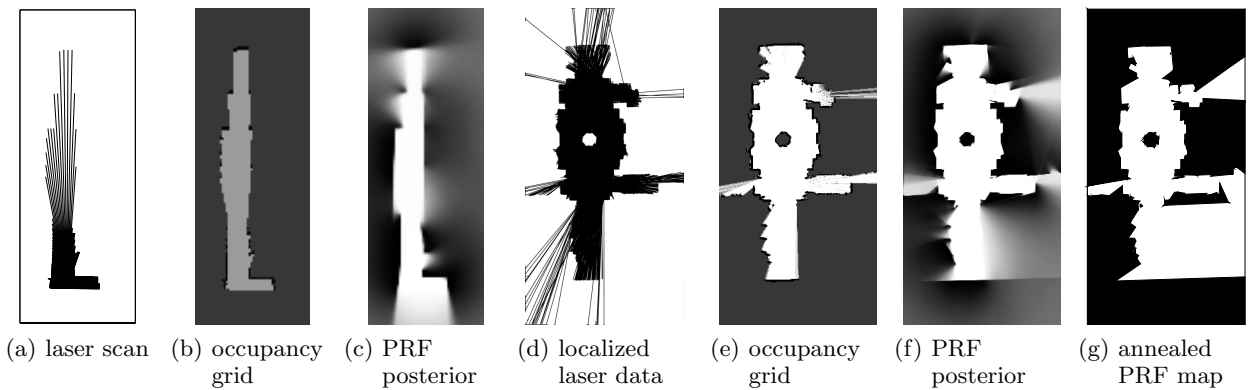

| (a) laser scan | (b) occupancy grid | (c) PRF posterior | (d) localized laser data | (e) occupancy grid | (f) PRF posterior | (g) annealed PRF map |

Figure 4: Experimental results with laser rangefinder data.

Once the return probabilities are computed for each feature, we compute the sonar likelihood using a mixture model. Each feature has a mixture component whose weight is its return probability and whose distribution is a Gaussian measurement error model centered at the feature's distance. There is also a mixture component representing an outlier model: its weight is the probability that none of the features generate a return, and its distribution is a mixture of a uniform distribution, an exponential distribution, and a point mass at the max-range measurement. This likelihood function is illustrated in Figure 3.

As with the laser data, we use a grid index over the sonar data to quickly identify which likelihoods must be recomputed when a local region of the coloring is updated. In the case of sonar, each observation is indexed in the cells that overlap a cone. Here too we use dynamic indexing to avoid unnecessary likelihood computations: when a sonar likelihood is recomputed, it is re-indexed in its cone up to the distance of its farthest contact (and to its maximum range if there are no contacts in a portion of its field of view).

## 3 EXPERIMENTAL RESULTS

We implemented the sampler described above in C++.[5] To investigate the sorts of maps that can be obtained from reasoning with PRFs, we collected a set of laser and sonar rangefinder data using a Pioneer 2DX robot and Player/Stage[6]. These observations were subsequently localized using the pmap library for laser-based localization and mapping.[7] In our experiments we found that the sampler is reasonably fast for smaller data sets: for example, on a Pentium 4 machine we obtain around 1,000 samples per second when processing a data set of 5,550 laser measurements; this leads to converged color estimates in under an hour. Our current implementation is slower when dealing with larger data sets; for our full sonar data set of 29,120 observations, convergence requires several hours.

In all experiments we set the scale parameter $p = 0.1$ (with length units in meters). This value was chosen by experimenting with several different values. As $p$ is increased beyond this range, structural complexity is penalized less, and surfaces become more jagged to fit errors in the data. When $p$ is decreased, structural complexity is penalized more and the posterior starts to smooth over interesting structures.

Perhaps the best way to illustrate the difference between the occupancy grid prior and the PRF prior is with a single laser scan. Figure 4(a) shows a single 180° laser scan taken from our data set; the robot is at the bottom of the figure, facing upwards. This scan was taken in a hallway, so there is a dense set of laser impacts near the robot, and sparse impacts farther down the hallway. Figure 4(b) shows an occupancy grid computed from this data: cells containing laser impacts are darker (more likely occupied) and cells that beams passed through are lighter; all other cells retain the prior occupancy probability of 50%.

---

[4] In (Tardós *et al.*, 2002) another sonar model is presented that also uses segment and corner features. Unlike our model, a single feature is chosen as the cause of each return, and the orientation of segments is used to deterministically rule out features where specular reflection would not return the wave to the sensor.

[5] http://paskin.org/prf
[6] http://playerstage.sourceforge.net
[7] http://robotics.usc.edu/~ahoward/pmap

Compare this occupancy grid to Figure 4(c), which visualizes the PRF posterior using a grid of point color estimates (as described above). It is immediately clear that the PRF prior yields significant extrapolation from the observations. First, the hallway is correctly classified as free space with high certainty. This occurs because of the spatial correlation in the PRF: each location's occupancy estimate is heavily influenced by nearby observations. We can also see how the PRF's bias towards linear color boundaries leads to useful extrapolation. In the laser scan, there are several subsets of collinear impacts. These provide strong evidence for planar boundaries that are hypothesized by the PRF; the end of the hallway (at the top) is a particularly good example because beyond the collinear impacts, the occupancy probability dwindles: the distribution is (correctly) unsure how far the wall extends.

This type of extrapolation also occurs in larger data sets. Figure 4(d) shows 5,550 laser observations, which represent $\frac{1}{10}$ of the laser data collected in the lobby of our building. Figure 4(e) shows the corresponding occupancy grid, and Figure 4(f) visualizes the PRF posterior. In regions where the data are unambiguous (as in the middle of the figure), the two maps largely agree. But where the data support multiple interpretations, the PRF posterior deviates in interesting ways. For example, we can see uncertainty in how far the lower wall extends; and, in the lower right quadrant, we can see an (uncertain) extrapolation of the free spaces. This type of extrapolation may prove useful in exploration tasks, where the robot must decide which part of the environment to observe next.

Using annealing, we can easily alter the sampler to *optimize* the map rather than average across different maps (Clifford and Nicholls, 1994). Figure 4(g) shows an optimized polygonal map obtained from the data in Figure 4(g). It picks out several minute details of the environment, including three doors. This map requires under 3KB to store with double precision coordinates; in contrast, the occupancy grid of Figure 4(e) requires 100KB with one byte per cell.

As mentioned above, it is difficult to improve upon occupancy grid techniques in the context of laser data because of the volume and quality of observations. This is not the case with sonar data. Figure 5(a) visualizes

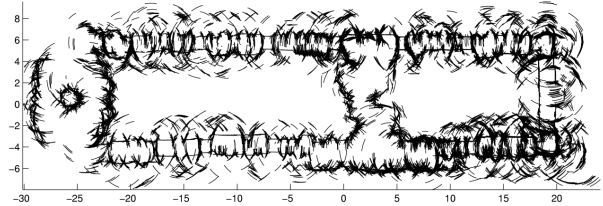
(a) localized sonar data (after max-range filtering)

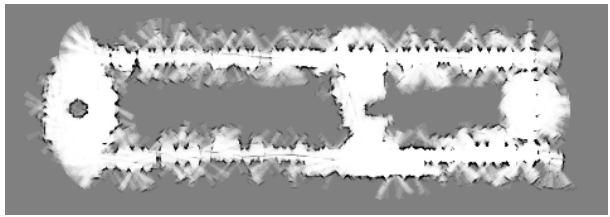
(b) the occupancy grid posterior

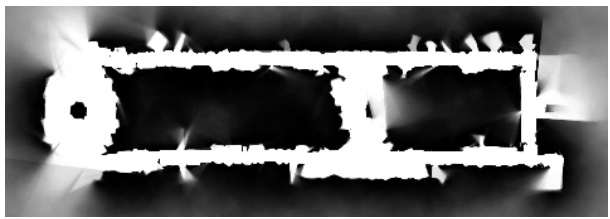
(c) the PRF posterior

Figure 5: The Gates Hall sonar data.

the impact arcs of our sonar data set (29,120 observations); we treated readings with depth greater than 3.5m as max-range measurements. Figure 5(b) shows the associated occupancy grid, which has indistinct (and sometimes missing) boundaries. In contrast, the PRF posterior, shown in 5(c), has crisp and accurate boundaries. Surprisingly, it also correctly identifies a number of rooms using a combination of subtle cues: a small number of longer range readings that enter the room, and corner impacts at the door frame.

## 4 DISCUSSION

Our experimental results demonstrate that polygonal random fields can yield significantly better maps than occupancy grids, especially on sonar data. These improvements arise for several reasons. PRFs have significant spatial correlation, which causes occupancy point estimates to be influenced by nearby observations. PRFs also have a bias towards linear boundaries between free and occupied space, which is helpful when mapping indoor environments. As a geometric representation, PRFs permit the rich sensor models required to interpret sonar data. We believe these advantages make PRFs a promising technique for probabilistic mapping, especially for ambiguous and noisy data. In this paper we have demonstrated improvements in sonar data, but there are other types of data that could benefit from more sophisticated inference,

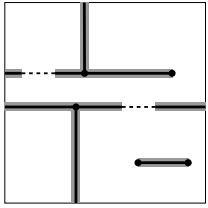

Figure 6: Illustration of a map "skeleton" and its induced coloring.

such as features extracted from stereo camera pairs.

However, there is much work to be done to make PRFs a practical alternative to current techniques. One key issue is space. Like other nonparametric probability distributions (e.g., Gaussian and Dirichlet processes), we must retain the data set to sample from the posterior. It would be interesting to explore techniques for summarizing the observed data, or techniques that fix a certain portion of the map—and forget the data used to build it—once its posterior is sufficiently peaked. Another possibility is to subsample the data and retain a small subset of observations necessary to define a peaked posterior.

Another important direction for future work is to extend the approach to perform simultaneous localization and mapping, so unlocalized measurements can be used. It is possible that particle filtering techniques, like those used in the FastSLAM (Montemerlo *et al.*, 2002) and DP-SLAM (Eliazar and Parr, 2004) algorithms, can be fruitfully applied in this context. Alternatively, the robot locations could be incorporated directly into the MCMC algorithm, allowing the sampler to jointly infer the trajectory and map.

Finally, while the pure Arak process we have used is a significantly better prior than occupancy grids, it not an ideal model of indoor environments. Buildings are mostly free space and are highly structured, unlike the samples in Figure 1. PRFs are quite flexible though, and the prior potential (3) can be augmented with cost functions that prefer discontinuities at certain orientations, maps with more empty space, etc. Another way to improve the prior is to perform inference over a restricted set of colorings. For example, to reason about indoor environments, we could use map "skeletons", like that shown in Figure 6, which represent walls by line segments; the actual coloring is represented implicitly, and is obtained explicitly by giving the walls thickness. The proposal distribution would operate directly on this lower-dimensional structure, but all prior and likelihood calculations would be based on the induced coloring. Techniques like this are likely to improve the mapping results, and provide additional useful extrapolation.

## Acknowledgements

We gratefully acknowledge Brian Gerkey for many helpful discussions and for assistance with gathering the laser and sonar data of Gates Hall. We also thank Geoff Nicholls for helpful discussions and for sharing his implementation of the C&N sampler, upon which our implementation was modeled.